\documentclass[sigconf]{acmart}

\usepackage{xcolor}

\AtBeginDocument{%
  }

\copyrightyear{2026}
\acmYear{2026}
\setcopyright{cc}
\setcctype{by}
\acmConference[HCOMP 2026]{2026 ACM Conference on Human-AI Complementarity and Alignment}{September 27--30, 2026}{Alexandria, VA, USA}
\acmBooktitle{2026 ACM Conference on Human-AI Complementarity and Alignment (HCOMP 2026), September 27--30, 2026, Alexandria, VA, USA}
\acmDOI{10.1145/3834580.3838751}
\acmISBN{979-8-4007-2894-5/2026/09}

\usepackage{tabularx}
\usepackage{array}
\usepackage{xspace}
\usepackage{algorithm}
\usepackage{algpseudocode}
\usepackage{amsmath}
\usepackage{booktabs}   
\usepackage{multirow}   
\usepackage{graphicx}   

\usepackage{pifont}     
\usepackage{xcolor}

\newcommand{\TReNN}{\textsc{TReNN}\xspace}
\newcommand{\ReNN}{\textsc{ReNN}\xspace}
\newcommand{\TNN}{\textsc{TNN}\xspace}
\newcommand{\SNN}{\textsc{SNN}\xspace}
\newcommand{\TEReNN}{\textsc{TE-ReNN}\xspace}
\usepackage{ulem} 

\newcommand{\best}[1]{\textbf{#1}}
\newcommand{\second}[1]{\underline{#1}}
\newcommand{\mstd}[2]{#1 {\scriptsize $\pm$ #2}}
\begin{document}

\title{Towards Actionable Surgical Team Dynamics: from Teamwork to Counterfactual Annotations}

 \author{Vincenzo Marco De Luca}
 \email{vincenzomarco.deluca@unitn.it}
 \orcid{0009-0003-3209-5817}
 \affiliation{%
   \institution{University of Trento}
   \city{Trento}
   \country{Italy}
 }

 \author{Antonio Longa}
 \email{antonio.longa@uit.no}
 \orcid{0000-0003-0337-1838}
 \affiliation{%
   \institution{UiT the Arctic University of Norway}
   \city{Tromsø}
   \country{Norway}
 }

 \author{Andrea Passerini}
 \email{andrea.passerini@unitn.it}
 \orcid{0000-0002-2765-5395}
 \affiliation{%
   \institution{University of Trento}
   \city{Trento}
   \country{Italy}
 }
\newcommand{\marco}[1]{\textcolor{red}{#1}}

\newtheorem{thm}{Theorem} 

\newtheorem{defn}[thm]{Definition} 


\begin{abstract}
Modeling team interactions in high-stakes environments such as surgical operating rooms is critical for understanding how coordination, communication, and individual behaviors jointly shape team performance and safety outcomes. However, existing multimodal datasets in this domain are often fragmented across modalities, annotation schemes, and formats, limiting their ability to support integrated analyses of real-world collaborative processes.

We address this limitation by introducing a curated and extended multimodal dataset for surgical team interaction analysis, built from real operating room recordings. Starting from an existing corpus, we construct a harmonized and analysis-ready version of the data by providing temporally consistent speaker diarization, bilingual (German and English) transcripts, and aligned multi-level annotations capturing team performance, interaction processes, and individual characteristics. Leveraging expert human annotations, team performance is assessed using a standardized surgical teamwork evaluation protocol, while interaction quality and individual attributes are annotated through structured rating schemes covering collaboration, group dynamics, and non-technical skills.

To further support the study of coordination breakdowns and performance variability, we introduce counterfactual annotations that describe plausible alternative team outcomes in the presence of observed interaction failures, enabling analysis of how specific behavioral patterns may relate to different trajectories of team performance. In addition, we provide structured temporal and relational representations designed to support computational modeling of teamwork processes and the design of AI-assisted collaborative systems.

Overall, the dataset is designed to support the study of how individual actions, interaction patterns, and team-level processes jointly contribute to team outcomes in surgical settings. It provides a unified resource for analyzing collaborative behavior in complex, real-world environments and enables both predictive modeling and behavioral analysis of team coordination in high-stakes domains.
\end{abstract}

\ccsdesc[500]{Human-centered computing~Collaborative and social computing theory, concepts and paradigms}
\ccsdesc[300]{Human-centered computing~Computer supported cooperative work}
\ccsdesc[100]{Applied computing~Health care information systems}
\keywords{Team analysis, Counterfactual explanations, Operating room teams, Surgical data science}

\maketitle

\section{Introduction}

Effective teamwork is essential for safety and performance in high-stakes collaborative environments~\cite{salas2005big} such as the operating room (OR)~\cite{haynes2009surgical}. In these settings, clinical outcomes depend not only on technical proficiency, but also on non-technical skills~\cite{flin2017safety}, including communication, coordination, leadership, situational awareness, and mutual support~\cite{yule2006non}. 
Failures in these dimensions can affect workflow and decision making~\cite{lingard2004communication}, ultimately compromising patient safety~\cite{moreal2025impact}.
These challenges have motivated growing interest in OR teamwork~\cite{weaver2010does}, training of surgical trainees~\cite{gaba2004future}, automatic tools for post-hoc assessment and real-time support~\cite{kiyasseh2023multi, ma2024artificial, kocielnik2024human, barba2026artificial}.

Recent progress in automatic team modeling has been supported by several team datasets~\cite{kraaij2005ami, sanchez2011audio, alameda2015salsa, beyan2016detecting, muller2018detecting}, but publicly available datasets for surgical environments remain limited~\cite{demir2022pocap, deng2025clinidial} or collect only a restricted set of modalities~\cite{srivastav2018mvor, demir2022pocap, deng2025clinidial}. 
This limitation is critical for team modeling since teamwork is inherently multimodal: it unfolds through speech, turn-taking, silence, interruptions, body conduct, role asymmetries, and temporally extended patterns of coordination~\cite{noel2022visualizing, luca2026boosting}. A dataset that only captures actions or procedural states provides an incomplete view of team interaction. 
Moreover, most existing work in this domain has focused on technical skill assessment~\cite{twinanda2016endonet, nwoye2020recognition, nwoye2022rendezvous} or scene understanding~\cite{ozsoy2025mm}, while less attention has been devoted to teamwork-centered computational models that characterize how surgical teams perform and evolve~\cite{seo2021towards,harari2024deep, kennedy2025novel}. 
Current approaches often reduce team performance to a single score, overlooking the complex interplay among several team processes.
Additionally, computational models designed for OR require interpretable accounts of teamwork to facilitate trust~\cite{dias2024clinician, subramanian2024designing, brandenburg2025can} and support suggestions when teamwork breakdowns occur~\cite{bates2003ten, endsley2017here, driskell2018teams}.

To address these limitations, we propose a new benchmark for surgical teamwork analysis built on top of a public OR dataset. The benchmark extends existing multimodal representations with speaker-aware and language-aware information, introduces a multi-level annotation framework for assessing teamwork, and provides a novel counterfactual annotation layer aimed at identifying events and behaviors associated with teamwork degradation.

In particular, the main contributions of this paper are summarized as follows: 

\begin{itemize}
\item Enrich multimodal feature extraction: hand-crafted speaker diarization, hand-crafted bilingual transcription
\item Multi-level annotation framework for the OR scenario: team-level, interaction-level, and individual-level assessment 
\item Counterfactual annotation framework for the OR scenario for identifying behaviors associated with teamwork degradation and collecting plausible suggestions to enhance team performance.
\end{itemize}


The remainder of the paper is organized as follows. Section~\ref{sec:related_work} reviews prior work on multimodal OR datasets, teamwork assessment frameworks, and socially grounded annotation resources. Section~\ref{sec:source_dataset} describes the source dataset, subset construction, and multimodal preprocessing pipeline. Section~\ref{sec:extended_mm_or} presents the annotation framework, including the multi-level teamwork labels and the proposed counterfactual annotation protocol, equipped with datasets and annotation quality analyses. 
Section~\ref{sec:benchmarks} introduces benchmark tasks and initial baselines. Finally, Section~\ref{sec:discussion} discusses limitations, release considerations, and future directions for multimodal teamwork modeling in high-stakes environments.

\section{Related Work}
\label{sec:related_work}
Prior work on modeling team behavior in clinical environments spans two main directions: multimodal OR datasets described in Sec.~\ref{subsec:operating_room_datasets} and team performance assessment frameworks analyzed in Sec.~\ref{subsec:team_performance_assessment}. These lines of research provide complementary insights but remain largely disconnected, limiting progress toward comprehensive modeling of team dynamics.

\subsection{Operating Room datasets}
\label{subsec:operating_room_datasets}

The development of multimodal OR datasets has enabled significant advances in surgical data science, particularly for perception and workflow understanding. Widely used benchmarks such as Cholec80 \cite{twinanda2016endonet}, Cholec40 \cite{nwoye2020recognition}, and Cholec50 \cite{nwoye2022rendezvous} provide annotated laparoscopic videos and have supported tasks such as surgical phase recognition and action detection. However, these datasets primarily focus on the visual content of the surgical field and do not capture team-level phenomena such as communication, coordination, or role dynamics. 

More recent efforts have explored multimodal clinical data beyond vision, in particular PoPCaP \cite{demir2022pocap} introduces a dataset combining speech, audio, and clinical imaging (i.e., chest X-rays), but lacks video recordings of real OR environments and is not publicly available, limiting both ecological validity and reproducibility. Similarly, CliniDial \cite{deng2025clinidial} provides clinical dialogue data suitable for natural language processing tasks, but excludes video due to privacy constraints, restricting the analysis to verbal interactions and preventing full multimodal modeling of embodied team behavior. Overall, existing multimodal OR resources either lack critical modalities, focus narrowly on surgical scenes rather than teams, or are not publicly accessible.

\textit{Our work aims to address this limitation by extending the MM-OR dataset to extract multimodal features that can encompass both operating room scenarios and team modeling settings, through rich data representations.}

\subsection{Assessing performance}
\label{subsec:team_performance_assessment}

In parallel, the medical and human factors literature has proposed structured frameworks for assessing teamwork, focusing on non-technical skills such as communication, leadership, coordination, and situational awareness. Instruments such as OTAS and NOTSS provide standardized rating schemes that are widely used for expert-based evaluation of team performance, offering strong theoretical grounding and validated constructs.

To bridge the gap between raw data and high-level constructs, recent work has explored socially grounded annotation resources that aim to capture interaction dynamics and team-level behaviors. These annotations provide a crucial link between multimodal signals and constructs such as coordination or shared situational awareness. 
However, such socially grounded annotations remain scarce, particularly in high-stakes environments such as the OR. Existing resources typically provide either low-level labels (e.g., actions and phases) or isolated modalities (e.g., dialogue transcripts, audio, or video), rather than jointly representing multimodal interactions and higher-level teamwork constructs. Moreover, privacy and ethical constraints make richly annotated OR data difficult to collect and share, further limiting the availability of resources suitable for computational learning of teamwork-related constructs.

\textit{Our work aims to address this gap in the literature by introducing multi-level annotations (i.e., team-level, individual-level, and interaction level) for teamwork and counterfactual annotations to make agent feedback accessible to be modeled and predicted through automatic learnable approaches.}

\section{Source Dataset}
\label{sec:source_dataset}

\begin{table}[t]
\centering
\begin{tabular}{l l}
\toprule
\textbf{Property} & \textbf{Description} \\
\midrule
Procedures & Robotic knee replacement \\
Recordings & 17 full-length + 22 short clips \\
Duration & $\sim$90 minutes per full procedure \\
Timepoints & 92,983 (1 FPS synchronized) \\
Data Volume & $\sim$500 GB \\
\midrule
Visual Modalities & 5 RGB-D, 3 RGB, 1 low-exposure RGB \\
Audio Modalities & 3 microphones \\
System Modalities & Robot logs, screen recordings \\
Tracking & Infrared 3D tracking \\
\midrule
Annotations & Segmentation, tracking, scene graphs \\
Downstream Tasks & Phase, next action, sterility breach \\
\bottomrule
\end{tabular}
\caption{Key characteristics of the MM-OR dataset.}
\label{tab:mmor}
\end{table}

Our work builds upon MM-OR, a publicly available multimodal OR dataset designed for the semantic understanding of complex surgical environments. MM-OR captures knee replacement procedures performed in a realistic OR setting using a medical-grade phantom and real clinical staff, enabling safe data collection while preserving realistic interaction dynamics.
The dataset includes 17 full-length surgical recordings (approximately 90 minutes each) and 22 short clips ranging from 1 to 180 minutes, for a total of 92,983 timepoints and approximately 500GB of multimodal data, further details are reported in Tab. \ref{tab:mmor}. Each timepoint corresponds to a temporally aligned snapshot across all modalities at 1 FPS. The dataset captures multiple surgical sessions across different days and team compositions, introducing variability in personnel, interactions, and workflow dynamics. This diversity makes MM-OR suitable for studying complex multi-agent behavior under realistic conditions. MM-OR provides a rich set of synchronized modalities capturing complementary aspects of the OR environment:

\begin{itemize}
\item Visual streams: five ceiling-mounted RGB-D cameras, three RGB cameras covering key areas, and one low-exposure RGB camera.
\item Audio and speech: recordings from three wireless microphones worn by surgical staff, capturing both verbal communication and tool-generated sounds, together with automatically generated time-aligned speech transcripts.
\item Robotic and system data: robotic logs, screen recordings, and procedural states describing system status.
\item 3D tracking: infrared marker-based tracking for tools, anatomy, and robotic components.
\end{itemize}

All modalities are hardware-synchronized, enabling fine-grained multimodal and temporal analysis of surgical workflows.

\paragraph{Annotations and Supported Tasks.}
MM-OR provides annotations for panoptic segmentation, object tracking, semantic scene graphs, and downstream tasks such as phase recognition, next-action anticipation, and sterility-breach detection.
Despite its multimodal richness, MM-OR primarily targets scene understanding and procedural modeling. Its annotations capture objects, actions, and their relationships, but not higher-level social and cognitive constructs such as communication quality, coordination, or non-technical skills.
In particular, MM-OR does not provide:
(i) speaker-aware conversational structure,
(ii) explicit modeling of teamwork dimensions,
or (iii) annotations describing behaviors perceived to contribute to teamwork breakdowns.

These limitations motivate our extension of MM-OR toward a teamwork-centered multimodal benchmark (Sec.~\ref{sec:extended_mm_or}).


\section{Extended MM-OR}
\label{sec:extended_mm_or}

This work extends MM-OR with an interaction-centric enrichment
and annotation layer comprising: (i) speaker diarization and
bilingual transcription; (ii) hierarchical teamwork annotations
spanning team, interaction, and individual levels; and (iii)
counterfactual annotations capturing behaviors perceived to
contribute to teamwork degradation. The resulting resource
combines multimodal observations with complementary human-
factors frameworks to support teamwork assessment and modeling.
In particular, the enriched multimodal data 
is described in Sec.~\ref{subsec:enrichment_pipeline}, 
the multi-layer annotation framework in Sec.~\ref{sec:annotation_framework}, and the novel counterfactual rating scheme 
in Sec.~\ref{subsec:annotation_counterfactual}.


\subsection{Multimodal Enrichment Pipeline}
\label{subsec:enrichment_pipeline}

To transform these data into a resource suitable for multimodal teamwork analysis, we applied an additional enrichment pipeline on top of the source data. The resulting resource combines the original multimodal signals with new hand-crafted speaker-aware and linguistically enriched representations that support annotation and future computational modeling. 


\paragraph{Speaker diarization.}
Speaker diarization identifies and segments speaker turns within each
clip. Given the noisy OR environment, diarization was manually refined
to ensure high temporal accuracy. This speaker-aware temporal
representation supports analyses of turn-taking, interruptions,
response latency, participation balance, and speaker-dependent
multimodal dynamics. In the present resource, diarization serves both
as an annotation aid and as a computational layer for downstream
modeling of conversational dynamics and team interaction patterns.

\paragraph{Bilingual transcripts.}
Each annotated clip is associated with bilingual transcripts in German
and English. The German transcripts preserve the original spoken
interaction, while the English translations facilitate annotation
review and computational analysis. The bilingual transcripts also
support future research on multilingual interaction and cross-lingual modeling of teamwork.

\subsection{Annotation Framework}
\label{sec:annotation_framework}

The rationale for multi-level teamwork annotations is that no single observational framework or global score can capture the full complexity of team dynamics in OR, which is a consequence of the dynamic interplay of communication, leadership, situational awareness, coordination, and role-dependent behavior over time. 

To address this limitation, we model teamwork through multiple complementary layers that capture distinct facets of team functioning, including team-level, interpersonal, and individual dimensions. A central implication of this design is that different levels of analysis may convey partially independent information, motivating a representation that explicitly separates them. In particular, Observational Teamwork Assessment for Surgery (OTAS)~\cite{undre2007observational} focuses on team-level performance, the Human-Machine Teaming (HMT) framework~\cite{meier2007rating} explores the interaction between pairs of team members, the General Leadership Impression Scale framework (GLIS)~\cite{lord1984test} and the System for the Multiple Level Observation of Groups (SYMLOG)~\cite{bales1980symlog, koenigs2002symlog} model analyze the Leadership, the Non-Technical Skills for Surgeons system (NOTSS)~\cite{yule2008surgeons, jung2018non} assesses non-technical skills of individual team members, and finally the Big-Five-In-Teamwork questionnaire (BFTQ)~\cite{van2012development} measures the teamwork quality of each individual team member.

\begin{table*}[t]
\centering
\renewcommand{\arraystretch}{1.4}
\begin{tabular}{
  >{\raggedright\arraybackslash}m{0.325\textwidth}
  | >{\raggedright\arraybackslash}m{0.625\textwidth}
}
\hline
\textbf{Term} & \textbf{Definition} \\
\hline
\textit{Communication} & Quality and quantity of information exchanged among members of
the team \\
\hline
\textit{Coordination} &  Management and timing of activities and tasks\\
\hline
\textit{Cooperation and Back-up behavior} & Assistance provided among members of the team, supporting others, and correcting errors \\
\hline
\textit{Leadership} & Provision of directions, assertiveness, and support among members of
the team \\
\hline
\textit{Team monitoring and Situational awareness} & Team observation and
awareness of ongoing processes \\
\hline
\end{tabular}
\caption{Definitions of team-level social construct introduced in OTAS} 
\label{tab:OTAS_definitions}
\end{table*}

\paragraph{Annotation setting}
The basic unit of annotation is a six-minute video clip, selected to provide sufficient temporal context for assessing teamwork while keeping the annotation process manageable across multiple frameworks. Each clip was independently reviewed by three annotators, who had access to the video, transcripts, and speaker-aware representations, and completed the full annotation protocol.  
This design allows us to capture both shared judgments and inter-rater variability across complementary teamwork frameworks, which is especially important given the complexity and partial subjectivity of the target constructs.
The protocol was designed to support both assessment and confidence scoring. To reduce ambiguity, annotators were instructed to base their decisions on the full temporal development. The procedure followed a consistent order across clips. Annotators first familiarized themselves with the segment and its interactional context. They then completed the framework-based teamwork annotations and finally provided the counterfactual judgments regarding the event or action that most negatively affected teamwork. This ordering was chosen deliberately: the descriptive ratings establish an overall view of the interaction, while the counterfactual layer encourages reflection on what may have driven the observed quality judgment.


\begin{figure}[t!]
    \centering
    \includegraphics[width=1\linewidth]{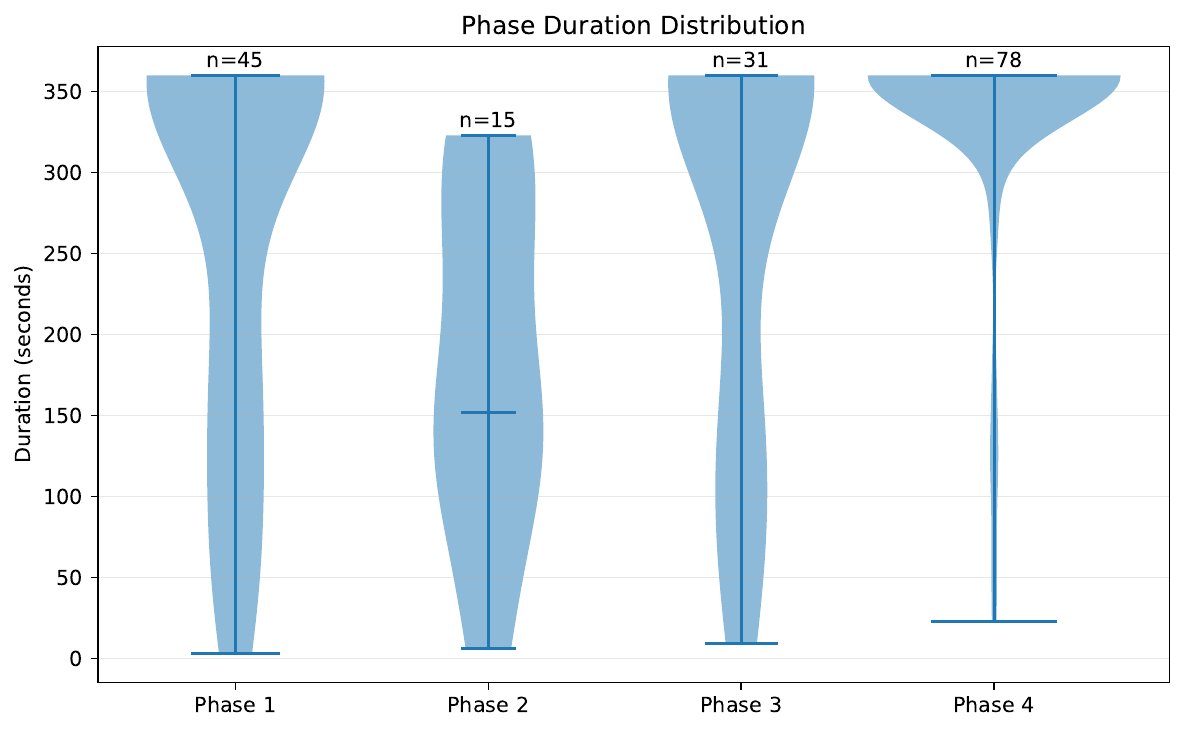}
    \caption{Violin plot of the distribution of samples across different phases, in particular we highlight the duration of each sample while $n$ represents the number of samples per each class.}
    \label{fig:phase_distributions}
\end{figure}

Each recording is partitioned according to the following events occurring in OR to guarantee coherence in the annotation stage:

\begin{itemize}
\item Phase 1: preliminary equipment stages: turn on procedure and the initial calibration
\item Phase 2: preliminary patient stages: dressing the robot to make it sterile, install the saw by nurse, and install base array by nurse
\item Phase 3: calibration stage: install calibration array, calibrate the robot by nurse, and remove calibration array
\item Phase 4: operation: install actual saw tip
\end{itemize}
In particular, the distribution of the data across the different phases and the distribution of the duration of each sample are reported in Fig. \ref{fig:phase_distributions}, in total we have 169 samples: 45 recorded in Phase 1, 15 in Phase 2, 31 in Phase 3, and 78 in Phase 4.
The phase partition choice is consistent with the intended use of the dataset extension: modeling how teams communicate, coordinate, and change their performance over time.

\begin{table*}[t]
\centering
\renewcommand{\arraystretch}{1.4}
\begin{tabular}{
  >{\raggedright\arraybackslash}m{0.28\textwidth}
  | >{\raggedright\arraybackslash}m{0.6\textwidth}
  | >{\centering\arraybackslash}m{0.07\textwidth}
}
\hline
\textbf{Team construct}  & \textbf{Definition} & \textbf{\# Items} \\
\hline
Communication
 & Maintaining mutual understanding among team members and managing the flow of dialogue to ensure continuity. & 2 \\
\hline
Joint information processing
 & Sharing information to optimize group decision-making and ability to reach an agreement among team members. &  2 \\
\hline
Coordination
 & Dividing tasks effectively among team members, managing time efficiently during teamwork, and coordinating technical skills to complete the task. & 3 \\
\hline
Interpersonal coordination
 & Promoting healthy interaction between team members to foster good relationships. & 1 \\
\hline
Motivation
 & Individual team member orientation towards completing individual tasks. &  1 \\
\hline
\end{tabular}
\caption{Definitions of interaction-level social constructs introduced in HMT}
\label{tab:HMT_definitions}
\end{table*}

\begin{table*}[t]
\centering
\renewcommand{\arraystretch}{1.4}
\begin{tabular}{
  >{\raggedright\arraybackslash}m{0.28\textwidth}
  | >{\raggedright\arraybackslash}m{0.6\textwidth}
  | >{\centering\arraybackslash}m{0.07\textwidth}
}
\hline
\textbf{Team construct}  & \textbf{Definition} & \textbf{\# Items} \\
\hline
 Situation Awareness	&  Gathering information from the environment and the team, correctly interpreting the information to understand the situation and anticipating future states and potential problems & 3 \\
    \hline 
    Decision Making  & Considering options and consequences before deciding, making appropriate decisions and communicating them clearly, implementing decisions and monitoring their outcomes & 3\\
    \hline
 Communication and Teamwork	& Sharing information clearly with team members, ensuring  team members understand tasks and plans, Organizing and coordinating tasks within the team & 3\\
 \hline

Leadership	& Encouraging and supporting team members, maintaining performance under stress, and	ensuring tasks progress efficiently and safely & 3\\
\hline
\end{tabular}
\caption{Definitions of individual social constructs introduced in NOTSS}
\label{tab:NOTSS_definitions}
\end{table*}

\begin{table*}[t]
\centering
\renewcommand{\arraystretch}{1.5}
\begin{tabular}{
  >{\raggedright\arraybackslash}m{0.28\textwidth}
  | >{\raggedright\arraybackslash}m{0.6\textwidth}
  | >{\centering\arraybackslash}m{0.07\textwidth}
}
\hline
\textbf{Team construct} & \textbf{Definition} & \textbf{\# Items} \\
\hline
\textit{Adaptability (A)} &
The ability of a team to adjust roles, strategies, and resource allocation in response to environmental changes~\cite{salas2005big}. & 5 \\
\hline
\textit{Back-up behavior (BB)} &
The ability of team members to anticipate the needs of others and provide timely support when required~\cite{salas2005big}. & 3 \\
\hline
\textit{Mutual performance monitoring (MPM)} &
The ability to comprehend the team environment and continuously monitor team members’ performance to ensure effective coordination~\cite{salas2005big}. & 5 \\
\hline
\textit{Team leadership (TL)} &
The capacity to direct, coordinate, and guide the activities of team members toward shared objectives~\cite{salas2005big}. & 2 \\
\hline
\textit{Team orientation (TO)} &
The capacity of team members to prioritize collective goals and outcomes over individual interests~\cite{salas2005big}. & 2\\
\hline
\hline
\textit{Closed-loop communication (CC)} &
The process by which team members explicitly direct, acknowledge, and confirm messages to ensure accurate receipt and shared understanding~\cite{salas2005big}. & 5\\
\hline
\textit{Mutual trust (MT)} &
The shared belief among team members that others will reliably fulfill their roles and act in the best interest of the team~\cite{salas2005big}. & 9\\
\hline
\textit{Shared mental models (SMM)} &
The common and overlapping understanding among team members regarding tasks, roles, and interaction patterns that supports effective coordination and anticipation of teammates’ actions~\cite{salas2005big}. & 4\\
\hline
\end{tabular}
\caption{Core behavioral components of the Big Five in Teamwork framework.}
\label{tab:main_tw_components}
\end{table*}

We organize the annotation schema into three levels of analysis: team-level performance, interaction-level assessment, and individual-level rating schemas.

\begin{figure}[t!]
    \centering
    \includegraphics[width=1\linewidth]{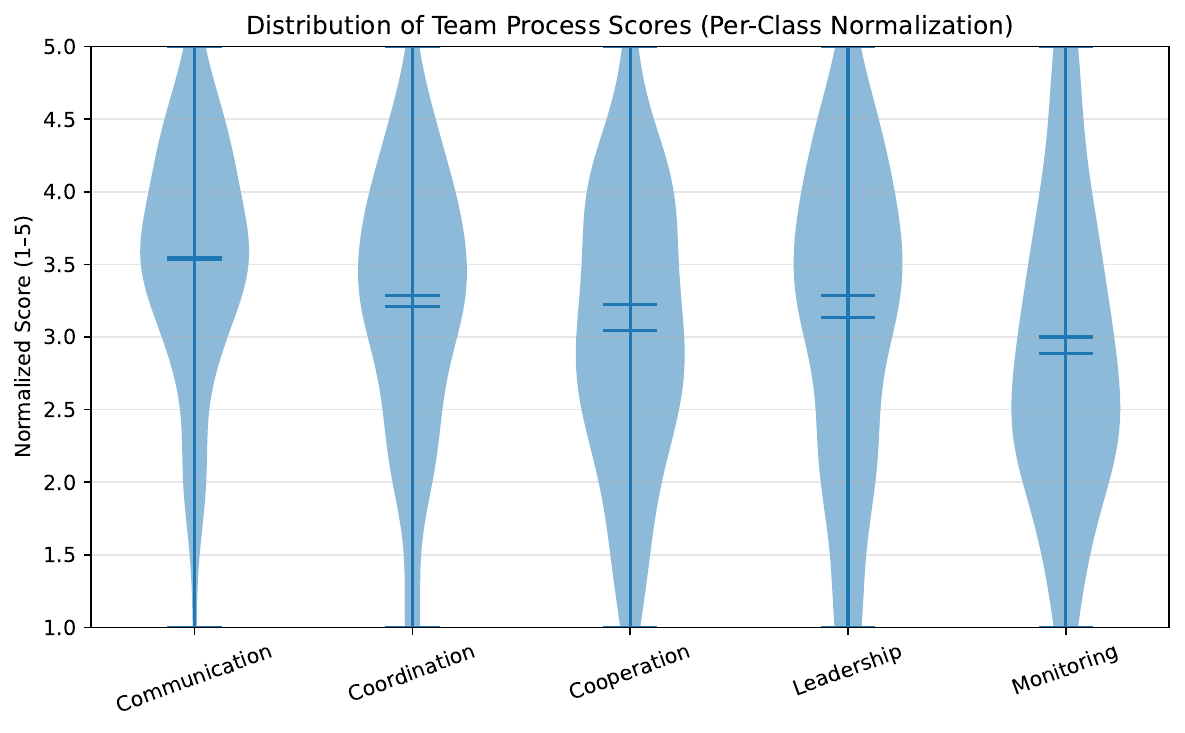}
    \caption{Violin plot of the distribution of samples for the OTAS rating schema, considering the average score provided by annotators and normalizing these scores in the range 1-5.}
    \label{fig:violin_otas}
\end{figure}

\paragraph{Team-level performance}
At the team level, we use the OTAS, which evaluates five core team constructs, each with a single item. Each construct is rated on a 0--6 Likert scale, where 0 indicates behavior that severely impairs team functioning, 3 indicates neutral performance, and 6 indicates exemplary teamwork.  The OTAS constructs capture core dimensions of teamwork grounded in established theory. Communication enables shared understanding and reduces errors~\cite{lingard2004communication, salas2005big}, while coordination structures the temporal alignment of interdependent tasks~\cite{undre2007observational}. Cooperation and backup behavior enhance team resilience and error recovery~\cite{salas2005big}. Leadership provides direction and regulates team interactions~\cite{yule2006non}. Finally, team monitoring and situational awareness support adaptive decision-making through a shared understanding of the evolving clinical context~\cite{endsley2017toward, salas2005big}. Additional details and definitions are provided in Table~\ref{tab:OTAS_definitions}.  The distribution of OTAS ratings, shown in Fig.~\ref{fig:violin_otas}, reveals a moderate class imbalance across all teamwork dimensions. Communication received predominantly positive evaluations, with 62.7\% of the clips assigned ratings 4 or 5, while only 16.0\% received ratings below 3. Coordination was concentrated around intermediate scores, with 42.7\% of the samples rated as 3, whereas ratings 1 and 5 each represented only 13.3\% of the observations. Cooperation followed a similar trend, with 65.3\% of the clips receiving ratings 3 or 4, while the highest rating (5) accounted for only 5.3\% of the samples. Leadership exhibited a more balanced distribution, although score 3 remained the most frequent class (34.7\%). Monitoring showed the lowest evaluations, with 46.6\% of the clips receiving ratings 1 or 2 and only 30.6\% receiving ratings 4 or 5. Overall, the annotations are concentrated around intermediate-to-positive scores, while extreme ratings are comparatively rare. This reflects the fact that the recorded surgical teams generally demonstrated acceptable teamwork performance, resulting in a realistic but moderately imbalanced label distribution.

\begin{figure}[t!]
    \centering
    \includegraphics[width=1\linewidth]{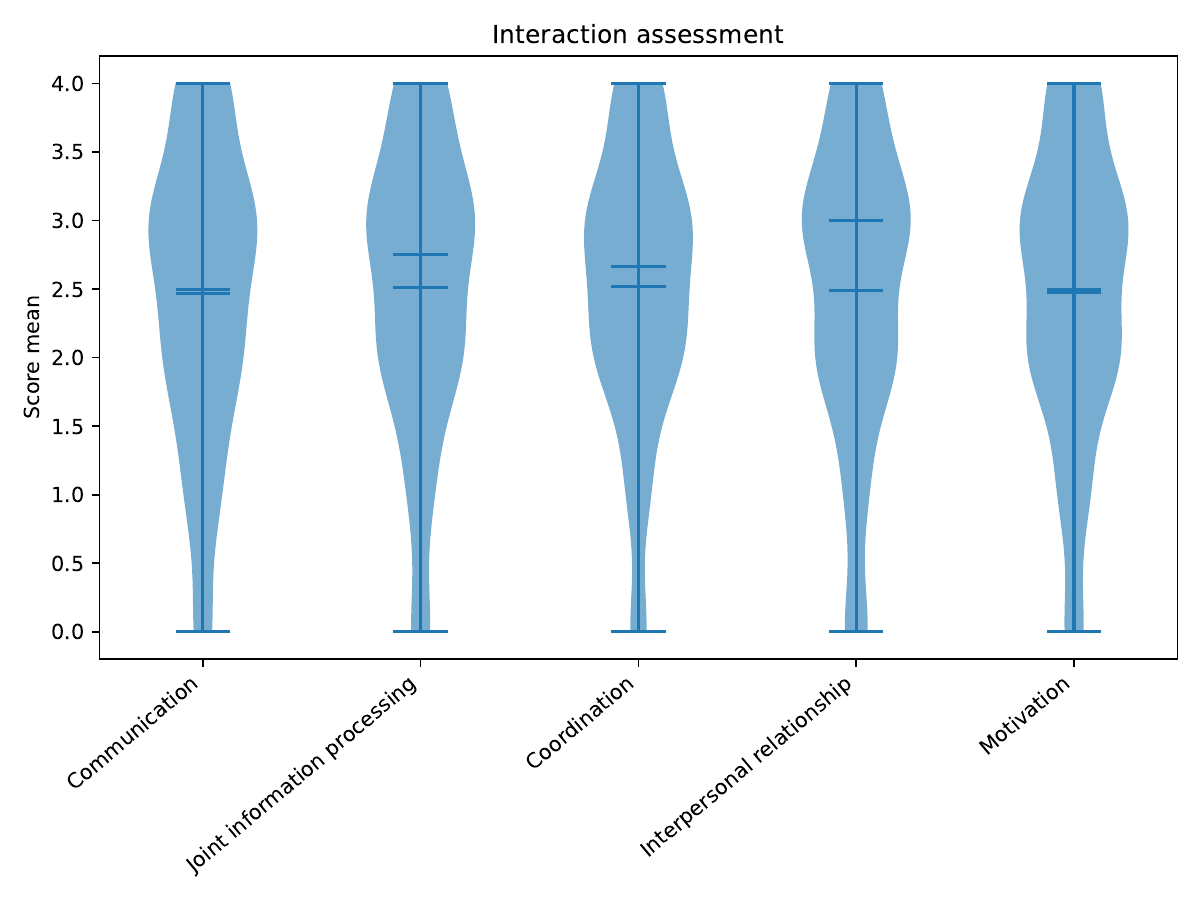}
    \caption{Violin plot of the distribution of samples for the HMT rating schema, considering the average score provided by annotators.}
    \label{fig:violin_hmt}
\end{figure}

\paragraph{Interaction-level assessment}
At the interaction level, we employ the HMT framework, adapted for human-only teams. HMT captures pairwise interaction quality across five macro-categories: communication (e.g., sustaining mutual understanding, dialogue management), joint information processing (e.g., information pooling, consensus building), coordination (e.g., task allocation and temporal coordination), interpersonal relationship (reciprocal interaction quality), and motivation (task engagement). These dimensions reflect core processes of effective teamwork in high-risk medical settings: communication and dialogue sustain shared mental models~\cite{Cooke2000, lingard2004communication}, joint information processing supports collective decision-making~\cite{salas2005big}, coordination structures interdependent surgical activities~\cite{undre2007observational}, interpersonal interaction fosters team cohesion~\cite{salas2005big}, and motivation captures individual engagement within team tasks~\cite{yule2006non}. Each category is scored on a 0--4 scale, with micro-level definitions provided in Table~\ref{tab:HMT_definitions}. Here, annotators labeled only relevant interactions, producing each of them more than 300 annotations. The distribution of HMT ratings is consistently centered around intermediate scores across all interaction dimensions, as shown in Fig.~\ref{fig:violin_hmt}. Classes 2 and 3 account for the majority of the annotations, representing between 63.6\% and 70.3\% of the samples. Communication and Joint Information Processing exhibit nearly identical distributions, with approximately one third of the observations assigned to class 2 and another third to class 3. Coordination shows the strongest concentration around intermediate ratings, with 70.3\% of the samples receiving class 2 or 3 labels. Low-quality interactions (classes 0 and 1) are relatively rare across all dimensions (12.3\%--13.8\%), while the highest rating (class 4) accounts for 17.4\%--22.6\% of the observations. Overall, the HMT annotations exhibit a moderate imbalance toward average-to-positive interaction quality while preserving sufficient variability to characterize both effective and suboptimal teamwork behaviors.

\begin{figure}[t!]
    \centering
    \includegraphics[width=1\linewidth]{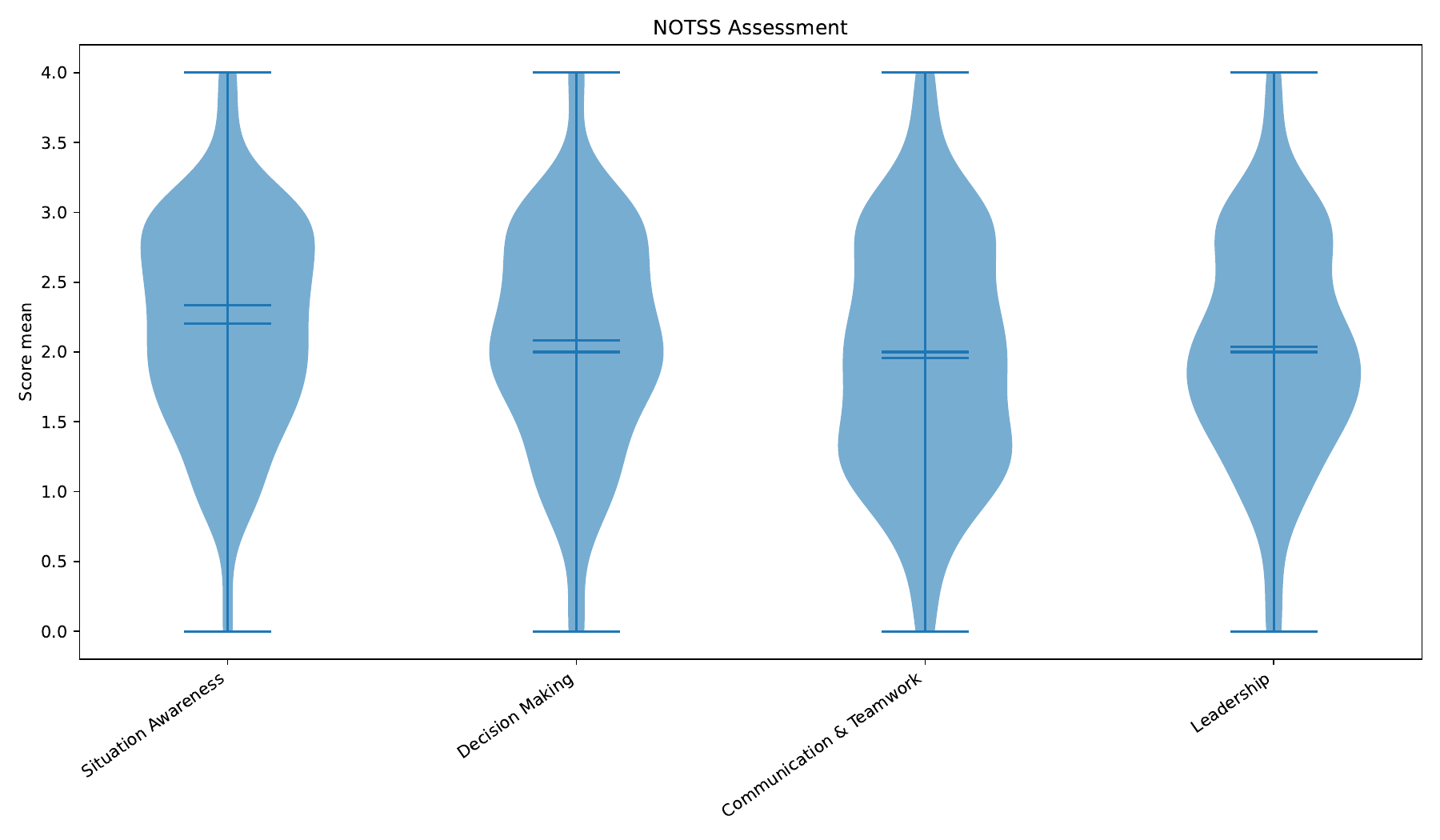}
    \caption{Violin plot of the distribution of samples for the NOTSS rating schema, considering the average score provided by annotators.}
    \label{fig:violin_notss}
\end{figure}

\paragraph{Individual-level schemas}
At the individual level, we distinguish between dynamic and static annotation schemas, reflecting different temporal sensitivities of behavioral constructs. As a dynamic framework, we adopt the NOTSS system, which captures rapidly evolving individual performance. NOTSS includes four macro-categories: communication and teamwork, decision making, leadership, and situational awareness, each operationalized through multiple items. These dimensions reflect core competencies for safe surgical practice: situational awareness supports accurate perception and anticipation of clinical states~\cite{endsley2017toward}, decision making enables timely and appropriate responses under uncertainty~\cite{flin2018development}, communication and teamwork ensure effective information exchange and coordination~\cite{lingard2004communication}, and leadership sustains team functioning and performance under stress~\cite{yule2006non}. Each category is scored on a predefined scale, with detailed definitions provided in Table~\ref{tab:NOTSS_definitions}.
The distribution of NOTSS ratings, shown in Fig.~\ref{fig:violin_notss} is predominantly concentrated around intermediate performance levels across all non-technical skill dimensions. Classes 2 and 3 account for the majority of the annotations, representing 64.7\%, 72.1\%, 57.4\%, and 66.7\% of the samples for Situation Awareness, Decision Making, Communication and Teamwork, and Leadership, respectively. Situation Awareness exhibits the most positive distribution, with 39.2\% of the samples receiving class 3 ratings, while Decision Making and Leadership are centered around class 2 (39.2\% and 40.7\%, respectively). Communication and Teamwork present the broadest distribution, with a larger proportion of lower ratings (37.7\% in classes 0 and 1) compared to the other dimensions. Across all constructs, extreme ratings are relatively uncommon, with class 0 accounting for only 2.5\%--7.8\% of the samples and class 4 for 3.4\%--4.9\%. Overall, the NOTSS annotations exhibit a moderate imbalance toward average performance levels, reflecting the realistic prevalence of competent but non-exceptional non-technical skills while preserving sufficient variability to characterize both suboptimal and highly effective behaviors.

\begin{figure}[t!]
    \centering
    \includegraphics[width=1\linewidth]{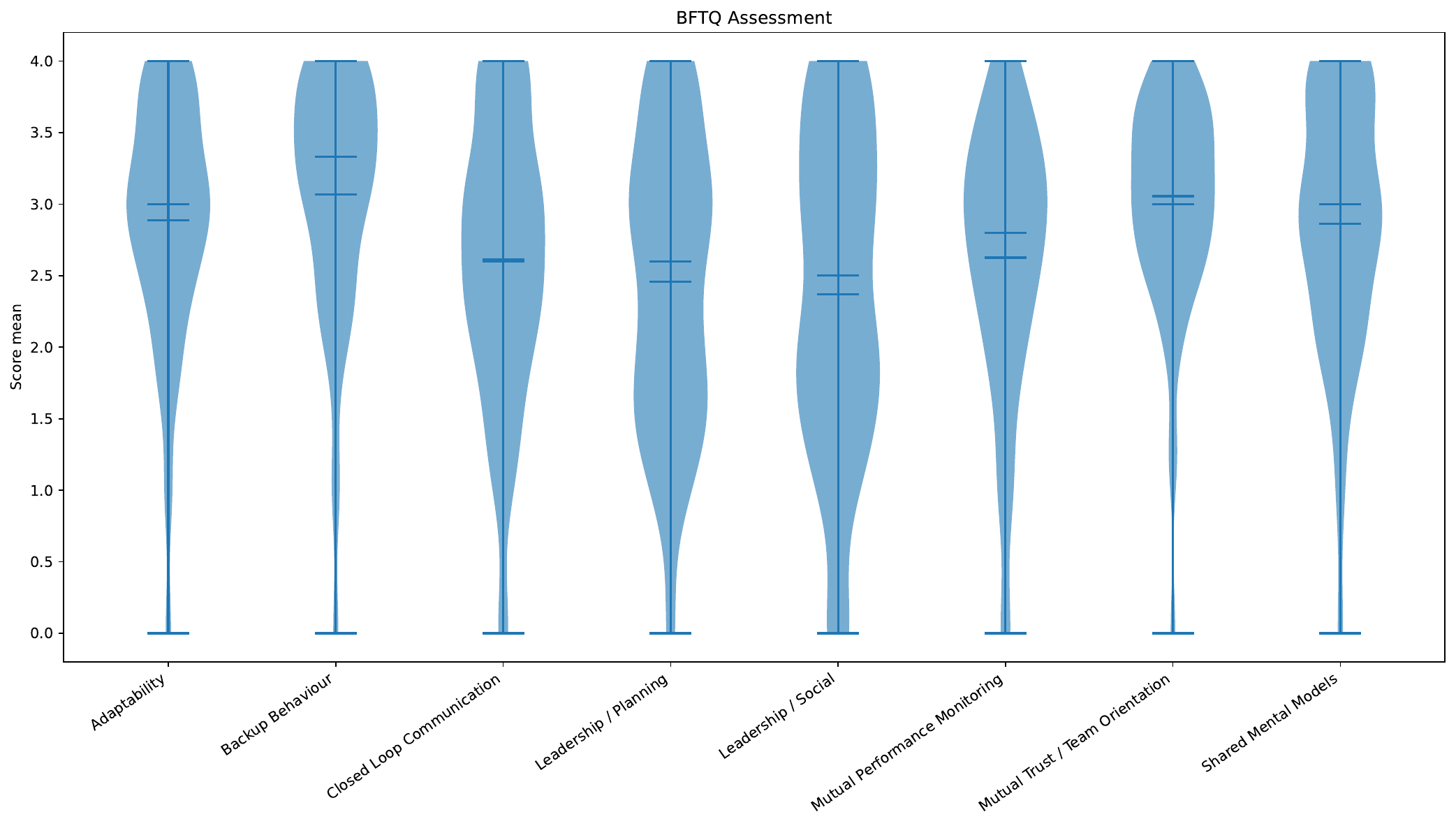}
    \caption{Violin plot of the distribution of samples for the BFTQ rating schema, considering the average score provided by annotators.}
    \label{fig:BFTQ}
\end{figure}

For more stable behavioral traits, we integrate three complementary frameworks. The BFTQ framework captures individual contributions to teamwork across multiple dimensions, including mutual trust/team orientation, leadership and planning, mutual performance monitoring, adaptability, closed-loop communication, social leadership, shared mental models, and back-up behavior. These constructs reflect well-established determinants of surgical team performance: mutual trust and team orientation enable coordinated action under high workload and uncertainty~\cite{salas2005big}; leadership and planning structure task execution in complex surgical environments~\cite{flin2018development}; mutual performance monitoring and back-up behavior support error detection and recovery~\cite{mcintyre1995measuring, marks2001temporally}, adaptability allows teams to respond effectively to unexpected intraoperative events~\cite{salas2005big}; closed-loop communication reduces information loss and improves patient safety~\cite{lingard2004communication}; social leadership fosters cohesion and coordination~\cite{yule2006non}, and shared mental models support aligned understanding of goals and procedures \cite{cannon1993shared, Cooke2000}. Each dimension is rated on a 0–4 scale, with further details provided in Table~\ref{tab:main_tw_components}. Here, annotators labeled only information depending on the agents active in the OR. Details about the distribution among classes are provided in Fig.~\ref{fig:BFTQ}.

\begin{figure}[t!]
    \centering
    \includegraphics[width=1\linewidth]{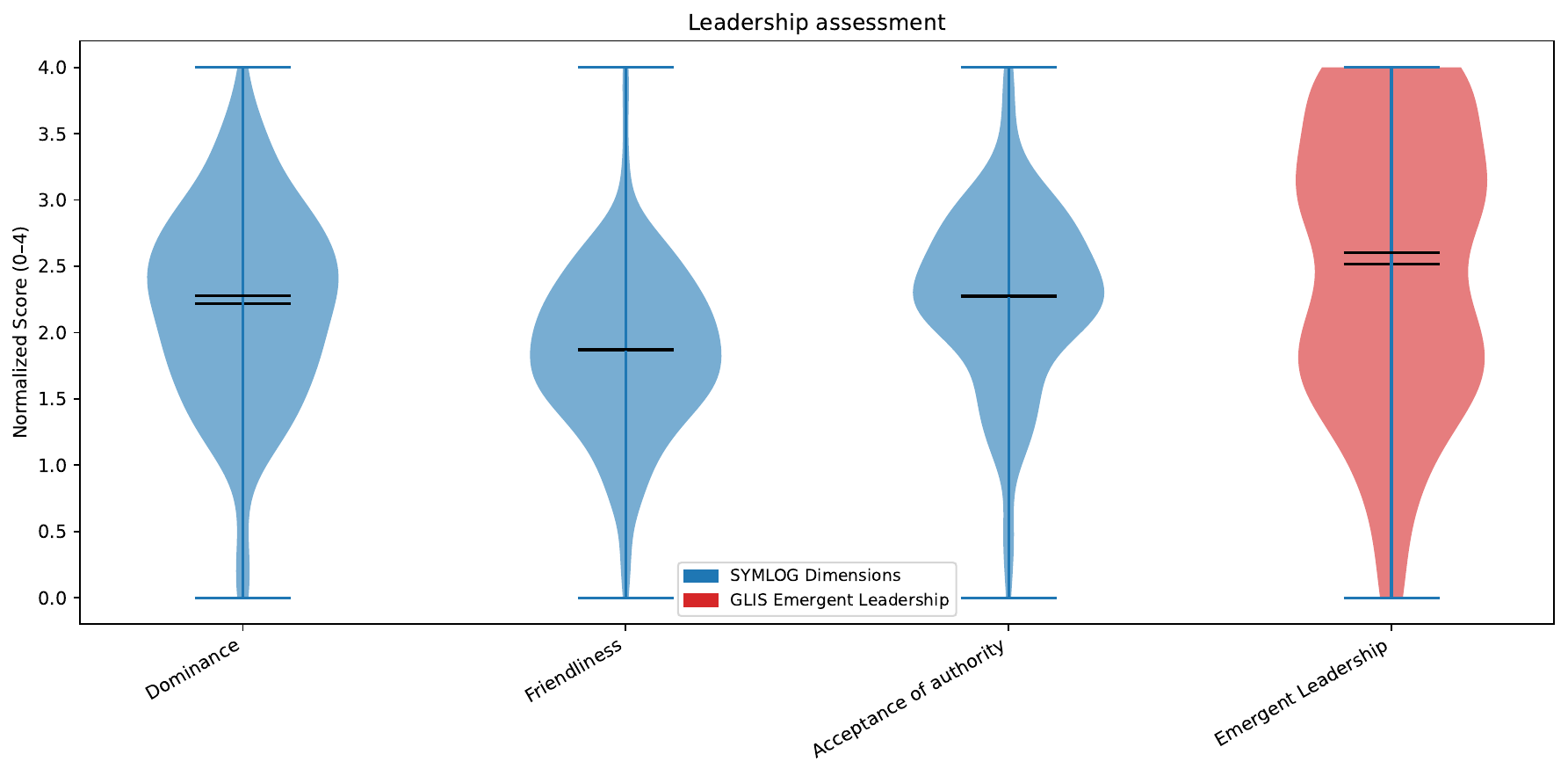}
    \caption{Violin plot of the distribution of samples for the SYMLOG and GLIS rating schemas, considering the average score provided by annotators.}
    \label{fig:leader_distribution}
\end{figure}

The SYMLOG model captures leadership and interpersonal behavior through 26 observational items structured along three bipolar dimensions: dominance vs. submissiveness, friendliness vs. hostility, and task-orientation vs. expressiveness~\cite{bales1980symlog}. Dominance reflects the extent to which individuals assume control, influence group direction, and shape decision processes; in surgical teams, such authority gradients are critical for coordinating action under time pressure, but excessive dominance may inhibit speaking-up behavior and degrade safety-critical communication~\cite{edmondson1999psychological, donaldson2000err, lingard2004communication}. Submissiveness captures reduced assertiveness and deference to authority, which can facilitate hierarchical efficiency but may increase the risk of unchallenged errors in the operating room~\cite{flin2017safety}. Friendliness vs. hostility represents the interpersonal climate of interaction, strongly linked to team psychological safety and communication openness, both of which are essential for error prevention in high-risk surgical environments~\cite{edmondson1999psychological, lingard2004communication}. Finally, task-orientation vs. expressiveness captures the balance between procedural focus and socio-emotional exchange, where strong task orientation supports procedural efficiency, while controlled socio-emotional interaction contributes to coordination and stress management during complex operations~\cite{salas2005big, yule2006non}. The distribution of SYMLOG ratings is strongly concentrated around intermediate behavioral profiles. Classes 2 and 3 account for 72.2\% of all observations, with class 2 representing the most frequent label (43.1\%). In contrast, extreme ratings are comparatively rare, with classes 0 and 4 accounting for only 3.6\% and 4.7\% of the samples, respectively. Lower ratings (classes 0 and 1) represent 23.0\% of the observations, while higher ratings (classes 3 and 4) account for 33.8\%. Overall, the distribution indicates a moderate imbalance toward average-to-positive interpersonal behaviors, reflecting the predominance of stable team interactions while preserving sufficient variability to characterize less effective and highly collaborative behaviors. Further details about the distribution of classes are provided in Fig. \ref{fig:leader_distribution}.

Finally, the GLIS framework contains five items capturing emergent leadership, defined as a dynamic behavioral state in which leadership influence arises during team interaction rather than being strictly tied to formal roles~\cite{stein1975identifying}. In this view, leadership is not a fixed attribute of a designated individual, but a context-dependent and temporally evolving phenomenon shaped by interactional dynamics and peer recognition. This is particularly relevant in the operating room, where leadership responsibilities may shift across team members depending on the phase of the procedure, the nature of the task, and the unfolding clinical situation. As a result, the same individual may assume or relinquish leadership influence multiple times within a single procedure. The distribution of GLIS ratings exhibits a moderate skew toward higher leadership effectiveness scores. While intermediate ratings remain common, with classes 2 and 3 accounting for 48.8\% of the samples, class 4 is the most frequent label, representing 32.4\% of all observations. In contrast, low leadership effectiveness ratings are comparatively rare, with classes 0 and 1 accounting for only 18.8\% of the samples. Overall, the distribution suggests that leadership interactions were generally perceived as effective, while still preserving sufficient variability to capture both suboptimal and highly effective leadership behaviors. The presence of observations across all five rating levels supports the use of GLIS for studying leadership quality under varying teamwork conditions. Further details about the distribution of ratings for GLIS are provided in Fig. \ref{fig:leader_distribution}.

At the core of the framework are annotations based on multiple established observational instruments; they were selected because they provide complementary views of surgical teamwork and non-technical performance. Rather than collapsing them into a unified latent score, we preserve them as distinct annotation layers. This allows the dataset to support different research questions, from team-level assessment to interpersonal analysis and cross-framework comparisons. More specifically, these frameworks capture overlapping but non-identical aspects of teamwork. Team-level frameworks emphasize collective coordination, communication, and cooperation. Individual-oriented frameworks capture non-technical skills such as leadership, awareness, and decision-making. Team-interaction frameworks make it possible to describe relational structure, behavioral positioning, or interpersonal climate. Taken together, these annotation layers reflect the view that teamwork quality is multi-dimensional, theoretically structured, and best analyzed through several concurrent lenses.

\subsection{Counterfactual Layer}
\label{subsec:annotation_counterfactual}


In addition to descriptive teamwork ratings, we introduce a counterfactual annotation layer aimed at capturing explanatory signals underlying team performance. For each six-minute clip, annotators identify the action, behavior, event, or interactional moment that, in their judgment, most negatively affected teamwork within the segment.

This layer addresses a limitation of standard teamwork annotations, which primarily characterize how well a team performs but provide limited insight into what drives performance degradation. Accordingly, it elicits structured judgments about perceived breakdown triggers, enabling a transition from purely descriptive to partially explanatory annotations. These triggers may include, for example, unclear directives, missed acknowledgments, delayed responses, or coordination and leadership failures.

\begin{figure}[t!]
    \centering
    \includegraphics[width=1\linewidth]{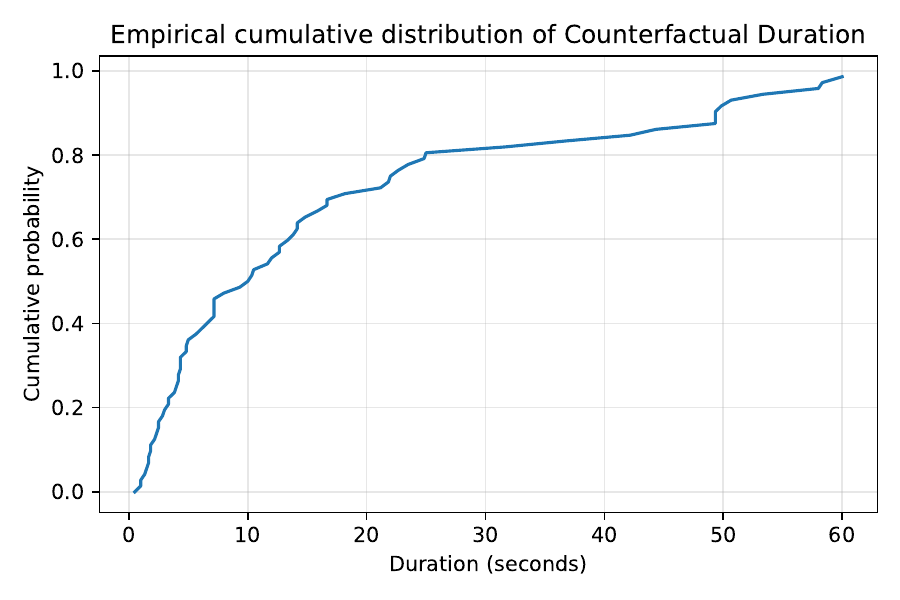}
    \caption{Empirical cumulative distribution function of the duration of the proposed counterfactual}
    \label{fig:counterfactual_duration}
\end{figure}

Each annotation is associated with a temporal interval $t$, an affected team construct $c$ selected from the OTAS dimensions (Communication, Coordination, Cooperation/Back-up behavior, Leadership, Monitoring/Situational Awareness), and at least one team member $a$ involved in the observed condition. This representation provides a structured yet flexible encoding of perceived breakdown events in team interaction data. The duration of these intervals is generally short (Fig. \ref{fig:counterfactual_duration}).

To further characterize the explanatory quality of each counterfactual annotation, we define a rating schema composed of five complementary dimensions, evaluated on an ordinal scale in $[1,5]$: criticality, plausibility, confidence, minimality, and expected enhancement. These dimensions are inspired by prior work on counterfactual reasoning and explainable AI, which emphasizes the importance of realistic, minimal, and actionable explanations for meaningful interpretation and intervention.

\textbf{Criticality} measures the impact of the identified condition on overall team performance, prioritizing events that materially affect task execution or outcomes.

\textbf{Plausibility} quantifies the extent to which the counterfactual represents a realistic and contextually feasible alternative, consistent with constraints imposed by the observed data and the context of the interaction~\cite{wachter2017counterfactual, karimi2020survey}.

\textbf{Confidence} captures the annotator’s certainty regarding the correctness of the annotation, including the identified condition, responsible agent(s), and affected construct, reflecting the importance of uncertainty modeling in human-generated explanations~\cite{miller2019explanation}.

\textbf{Minimality} evaluates whether the counterfactual corresponds to the smallest sufficient modification required to alter the observed condition, aligning with the principle that informative explanations should isolate the relevant behavioral factor without introducing unnecessary changes~\cite{wachter2017counterfactual, karimi2020survey}.

\textbf{Expected enhancement} measures the extent to which the proposed counterfactual is expected to improve team performance, capturing the action-oriented nature of counterfactual reasoning in human-centered AI systems~\cite{miller2019explanation}.

Collectively, these dimensions ensure that counterfactual annotations are impact-driven, plausible, reliable, minimally sufficient, and actionable, supporting both interpretability and downstream modeling of complex team interactions.
We additionally collected textual rationales describing what happened, its consequences, why it should be avoided, and how team members could behave differently from how they actually behaved.
Counterfactual annotations are primarily associated with communication-related failures (approximately 28.0\% of all events), followed by monitoring and situational awareness issues (22.7\%) and cooperation or backup behavior breakdowns (20.0\%). In contrast, leadership- and coordination-related counterfactuals are less frequent, each accounting for approximately 12.0\% of the annotations. Overall, nearly 70\% of all counterfactuals concern communication, monitoring, and cooperation processes, highlighting these dimensions as the most common sources of teamwork breakdowns in the observed surgical procedures.
Most counterfactuals identify a single responsible actor (mean = 1.3 actors per event), suggesting that annotators generally attributed failures to specific individual actions rather than collective team behavior. Responsibility is most frequently assigned to actors ID2 and ID1, while the remaining team members appear substantially less often. Regarding counterfactual quality, the score distributions reveal distinct characteristics across the five dimensions. Criticality is strongly skewed toward lower values, with 65.3\% of the events receiving scores 1 or 2 and no event assigned the maximum score of 5, indicating that most counterfactuals concern moderate teamwork inefficiencies rather than severe failures. Confidence follows a similar pattern, with 66.7\% of the annotations rated 1 or 2, suggesting that annotators often recognized alternative interpretations of the observed situations. In contrast, plausibility is concentrated around intermediate values, with 73.3\% of the counterfactuals receiving scores 2 or 3, indicating that the proposed interventions are generally realistic and achievable. Minimality exhibits the most favorable distribution, with 70.7\% of the annotations rated 3 or higher and 44.0\% receiving scores 4 or 5, suggesting that most counterfactuals involve relatively small behavioral modifications. Finally, Performance Gain is centered on intermediate values, with 54.7\% of the annotations assigned scores 2 or 3 and only 25.3\% receiving scores 4 or 5, indicating that the proposed interventions are expected to provide incremental rather than transformative improvements in team performance. These annotations should be interpreted as expert judgments about plausible alternative behaviors rather than as experimentally validated causal effects.

\begin{figure}[t!]
    \centering
    \includegraphics[width=1\linewidth]{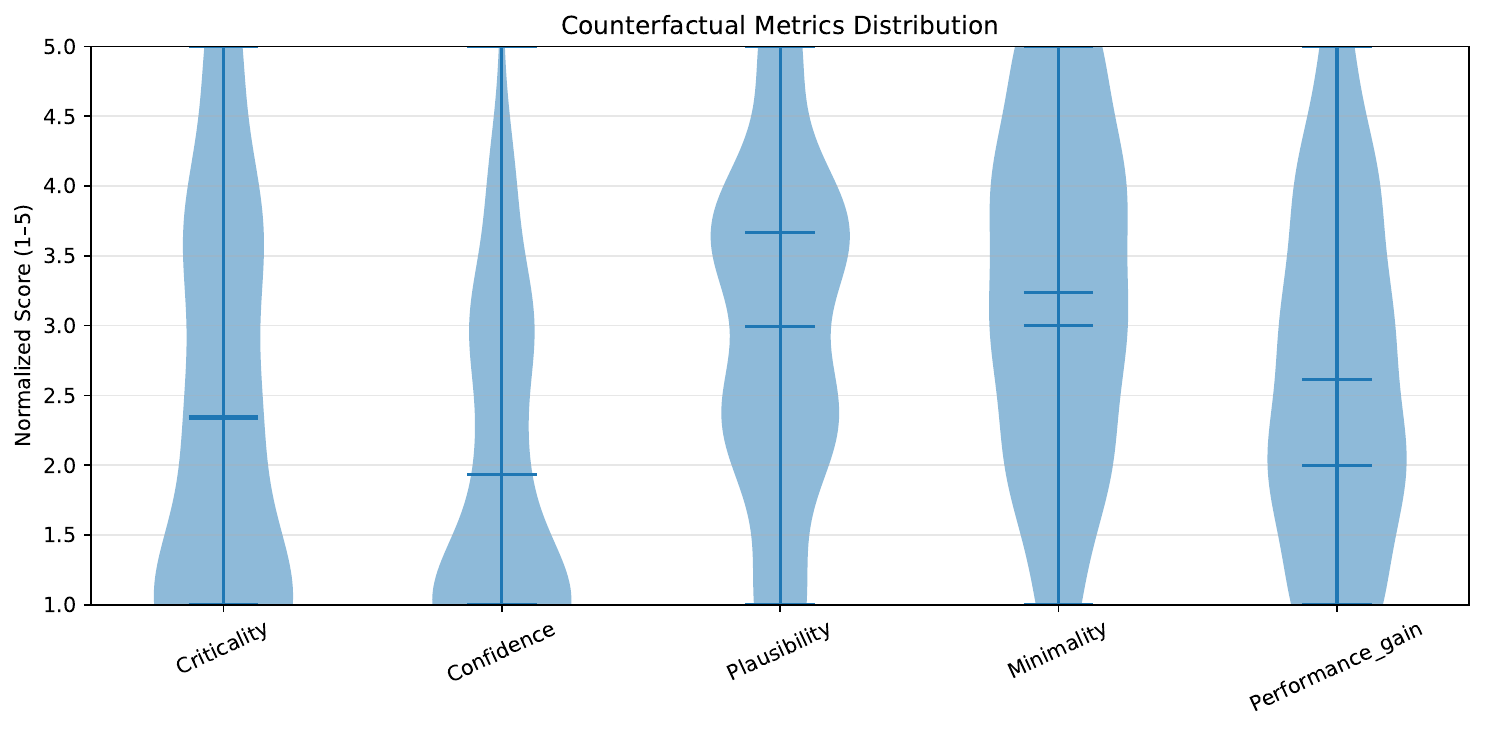}
    \caption{Violin plot of the distribution of the measures provided for the proposed counterfactual rating schemas.}
    \label{fig:counterfactual_metrics}
\end{figure}

\section{Benchmarks}\label{sec:benchmarks}
The proposed extension is designed not only as a collection of annotations, but as a resource for studying teamwork in operating rooms from multiple complementary perspectives. To illustrate its potential impact, we present a set of benchmark tasks that demonstrate how the enriched multimodal representations, the multi-level annotation framework, and the counterfactual layer can support different lines of future research.
Specifically, we focus on three representative use cases: the study of multimodal feature enrichment and conversational representations for teamwork modeling in Sec. \ref{subsec:feature_multimodal},
the prediction of teamwork-related constructs at team, interactional, and individual levels in Sec. \ref{subsec:predictive_power}, and 
the automatic identification of counterfactual events associated with teamwork degradation in Sec. \ref{subsec:counterfactual_prediction}.
These experiments are not intended to establish state-of-the-art performance, but rather to provide initial baselines and demonstrate the feasibility and applicability of the proposed dataset for future multimodal teamwork research.

\subsection{Feature enrichment pipeline}
\label{subsec:feature_multimodal}
\begin{figure}[h!]
    \centering
    \includegraphics[width=1\linewidth]{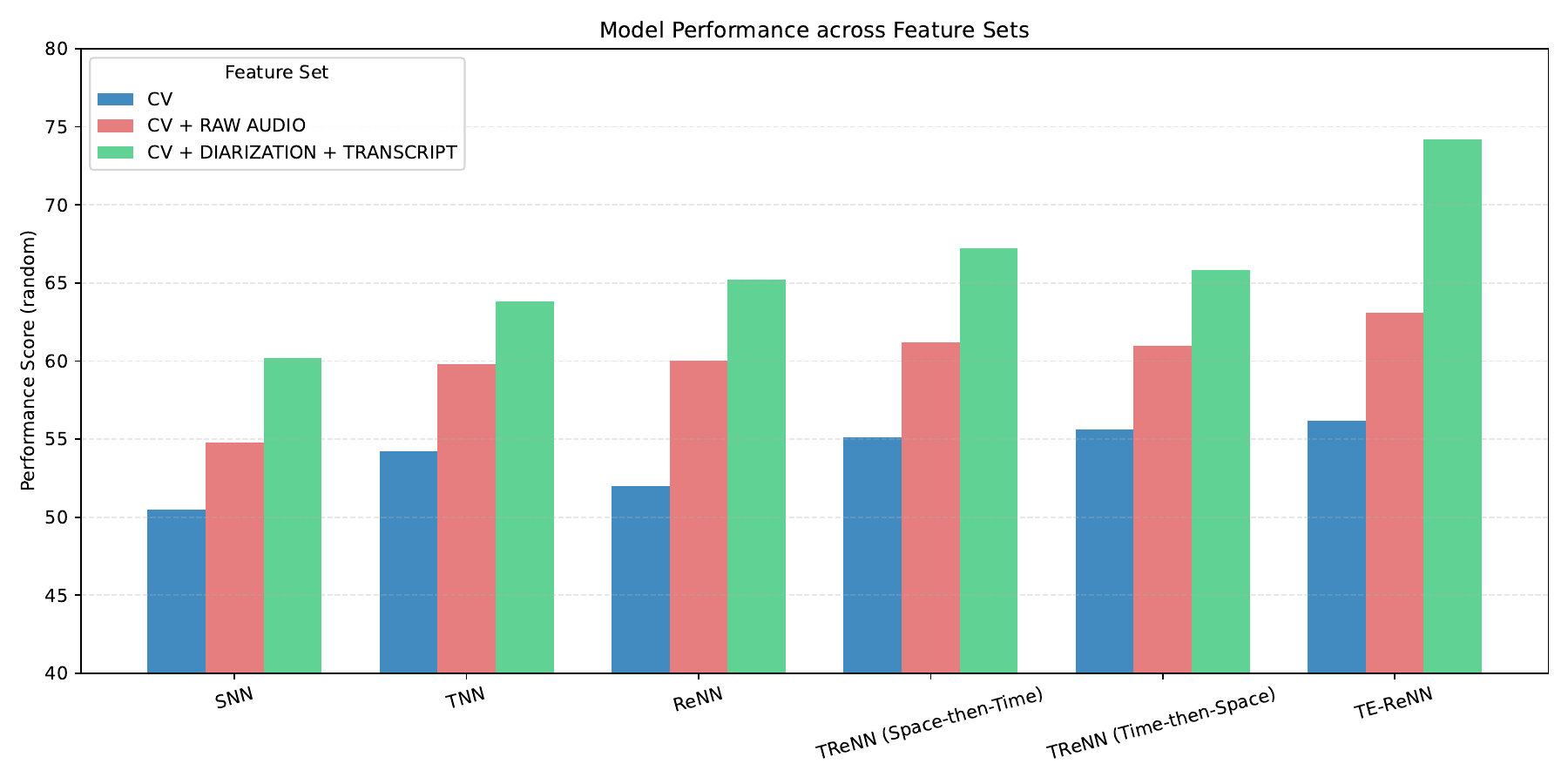}
    \caption{Average F1-macro over the different OTAS dimensions. For each modeling paradigm, the results reported correspond the best-performing model.}
    \label{fig:multimodal_otas}
\end{figure}

One of the contributions of this work is the enrichment of MM-OR with speaker-aware and language-aware representations. While these additional annotation layers are designed to facilitate the study of surgical teamwork, to prove its practical value, we analyze the impact of the proposed multimodal enrichment on downstream teamwork modeling.
Specifically, we progressively augment the original MM-OR representation with additional sources of information and compare the resulting predictive performance. The considered feature configurations include:  structural relational information available in the original dataset,  computer-vision-derived features such as motion patterns and interpersonal distances, paralinguistic audio descriptors, and textual embeddings and relational information extracted through diarized transcripts.
The goal of this analysis is not to establish a new state of the art, but rather to quantify the utility of the enrichment pipeline and to assess whether speaker-aware conversational representations provide complementary information beyond raw multimodal signals. Previous work has shown that structural and temporal conversation context can improve text classification, although its benefits depend on dataset size and interaction complexity \cite{penzo2024putting}. We therefore empirically assess, rather than assume, the value of transcript-derived relational context for teamwork modeling.
To provide a comprehensive evaluation, we consider multiple modeling paradigms characterized by different inductive biases, following the taxonomy proposed in~\cite{luca2026boosting}. In particular, we evaluate: (i) static feature-based models (\SNN), which process each observation independently; (ii) temporal models (\TNN), which capture behavioral evolution over time; (iii) relational models (\ReNN), which explicitly model interactions among team members; and (iv) hybrid spatio-temporal architectures (\TReNN and \TEReNN~\cite{de2026actionable}), which jointly reason over temporal and relational dependencies~\cite{longa2023graph}.
To ensure a fair comparison across feature configurations, in the "Raw-audio" configuration where diarization is assumed not to be available, audio is represented through global acoustic embeddings. Conversely, the enriched setting incorporates speaker-level paralinguistic descriptors together with textual representations extracted from the manually curated transcripts. For relational architectures, interaction graphs are constructed using spatial proximity, while diarization additionally enables a speaker-centric communication topology in which the active speaker is connected to the remaining team members, following prior work~\cite{luca2026boosting}.
Figure~\ref{fig:multimodal_otas} summarizes the results obtained across feature configurations and modeling paradigms. For clarity, we report the best-performing architecture within each family: Random Forest (RF) for \SNN, Multi-Head Attention (MHA) for \TNN, Graph Isomorphism Networks (GIN) for \ReNN, the best-performing GNN+MHA variants for \TReNN, and TE-GIN for \TEReNN. All experiments follow a Leave-One-Group-Out (LOGO) protocol, where surgical teams are held out during testing to evaluate generalization across unseen team compositions. All clips originating from the same surgical procedure were assigned to the same fold, preventing clips from a given surgical procedure from appearing in both training and test sets. Experiments are averaged over ten runs with different random seeds.
Overall, the results demonstrate that the proposed enrichment pipeline consistently improves teamwork prediction performance. In particular, speaker-aware and transcript-based representations provide substantial gains over raw audio features, suggesting that conversational structure constitutes a critical component for modeling collaboration in operating room environments. These findings support the relevance of the additional annotation effort and highlight the potential of the proposed resource for future multimodal teamwork research.

\subsection{Multi-level prediction benchmark}
\label{subsec:predictive_power}

\begin{table*}[h!]
\centering
\resizebox{\textwidth}{!}{
\begin{tabular}{l | l || c c || c c c c || c}
\toprule
\multirow{2}{*}{Paradigm} & \multirow{2}{*}{Model}
& \multicolumn{2}{c||}{Team}
& \multicolumn{4}{c||}{Individual}
& Interaction \\
\cmidrule(lr){3-4}
\cmidrule(lr){5-8}
\cmidrule(lr){9-9}
&
& OTAS
& Phase duration
& NOTSS & BFTQ & SYMLOG & GLIS
& HMT \\
\midrule

\multirow{2}{*}{\SNN}
& MLP   
& \mstd{59.8}{2.2} & \mstd{60.5}{1.2} & \mstd{51.5}{1.0} & \mstd{42.5}{1.1} & \mstd{48.5}{0.9} & \mstd{64.5}{0.8} & \mstd{41.5}{1.2} \\
& RF    
& \mstd{60.2}{0.8} & \mstd{61.8}{1.8} & \mstd{52.1}{1.2} & \mstd{38.5}{0.7} & \mstd{42.1}{0.6} & \mstd{65.2}{0.7} & \mstd{44.1}{1.0} \\

\midrule
\multirow{3}{*}{\TNN}
& LSTM  
& \mstd{62.3}{1.6} & \mstd{62.8}{0.9} & \mstd{56.2}{1.4} & \mstd{52.2}{1.3} & \mstd{55.2}{0.9} & \mstd{64.8}{1.1} & \mstd{48.2}{1.1}  \\
& GRU   
& \mstd{61.8}{1.8} & \mstd{63.0}{0.8} & \mstd{58.1}{1.1} & \mstd{55.5}{1.2} & \mstd{57.3}{0.8} & \mstd{62.6}{1.3} & \mstd{51.2}{1.3} \\
& MHA   
& \mstd{63.8}{2.0} & \mstd{62.4}{1.3} & \mstd{60.4}{1.2} & \mstd{56.2}{1.0} & \mstd{61.3}{0.8} & \mstd{64.9}{1.0} & \mstd{53.0}{1.2}  \\

\midrule
\multirow{3}{*}{\ReNN}
& GCN   
& \mstd{61.4}{1.5} & \mstd{64.6}{0.6} & \mstd{58.2}{1.5} & \mstd{57.5}{1.1} & \mstd{59.8}{0.9} & \mstd{63.7}{1.2} & \mstd{56.2}{1.2}   \\
& GAT   
& \mstd{62.1}{1.8} & \mstd{65.2}{0.8} & \mstd{59.3}{1.0} & \mstd{58.5}{1.2} & \mstd{59.6}{1.0} & \mstd{65.4}{1.2} & \mstd{58.0}{1.2} \\
& GIN   
& \mstd{65.2}{1.5} & \mstd{64.0}{1.2}  & \mstd{61.2}{1.1} & \mstd{59.0}{1.3} & \mstd{60.4}{1.1} & \mstd{66.8}{1.0} & \mstd{58.3}{1.1} \\

\midrule
\multirow{3}{*}{\shortstack[l]{Space-and-Time\\\TReNN}}
& GCN+MHA 
& \mstd{64.0}{1.5} & \mstd{67.3}{0.6} & \mstd{66.6}{1.1} & \mstd{62.8}{1.0} & \mstd{64.1}{0.8} & \mstd{62.8}{1.2} & \mstd{65.4}{1.6} \\
& GAT+MHA 
& \mstd{67.2}{1.7} & \mstd{67.1}{0.7}  & \mstd{65.6}{1.1} & \mstd{65.8}{1.2} & \mstd{66.1}{1.1} & \mstd{62.0}{0.8} & \mstd{65.1}{1.2}  \\
& GIN+MHA 
& \mstd{63.2}{1.5} & \mstd{65.2}{0.6} & \mstd{66.2}{1.0} & \mstd{65.2}{1.1} & \mstd{66.0}{1.2} & \mstd{64.2}{1.0} & \mstd{65.2}{1.3} \\

\midrule
\multirow{3}{*}{\shortstack[l]{Time-and-Space\\\TReNN}}
& MHA+GCN 
& \mstd{65.2}{1.3} & \mstd{64.8}{0.8} & \mstd{64.0}{1.2} & \mstd{65.2}{1.8} & \mstd{63.8}{1.0} & \mstd{65.2}{0.9} & \mstd{69.4}{1.1}\\
& MHA+GAT 
& \mstd{65.8}{1.5} & \mstd{64.8}{1.0} & \mstd{64.1}{1.0} & \mstd{65.1}{1.6} & \mstd{64.0}{1.2} & \mstd{67.6}{0.9} & \mstd{66.8}{1.1} \\
& MHA+GIN 
& \mstd{64.2}{1.2} & \mstd{65.6}{0.9} & \mstd{65.0}{1.2} & \mstd{64.1}{1.4} & \mstd{63.2}{1.1} & \mstd{68.4}{1.0} & \mstd{67.9}{1.0} \\

\midrule
\multirow{3}{*}{\TEReNN}
& TE-GCN 
& \mstd{72.4}{1.7} & \second{\mstd{70.1}{1.0}} & \mstd{71.0}{1.3} & \mstd{76.1}{1.4} & \mstd{70.2}{1.1} & \mstd{75.4}{1.0} & \mstd{73.9}{1.2} \\
& TE-GAT 
& \second{\mstd{73.1}{1.7}} & \mstd{69.8}{0.8} & \mstd{71.7}{1.2} & \mstd{75.5}{1.1} & \mstd{71.5}{1.1} & \mstd{74.8}{1.0} & \best{\mstd{73.9}{0.8}}  \\
& TE-GIN 
& \best{\mstd{74.2}{1.7}} & \best{\mstd{72.5}{1.0}} & \best{\mstd{72.5}{1.0}} & \mstd{76.7}{1.1} & \mstd{73.9}{1.0} & \best{\mstd{76.2}{1.2}} & \mstd{73.5}{0.8}  \\

\bottomrule
\end{tabular}
}
\caption{Combined results measuring F1-macro according to a LOGO protocol, with experiments run over 10 different random seeds. Tasks are organized into three levels: Team (OTAS, Phase duration), Individual (NOTSS, BFTQ, SYMLOG, GLIS), and Interaction (HMT). Bold indicates best, underline indicates second best.}
\label{tab:combined_results}
\end{table*}

The proposed dataset extension introduces a collection of complementary annotation frameworks capturing different aspects of teamwork. Unlike existing operating-room benchmarks, which primarily focus on workflow recognition or procedural understanding, our extension enables the study of teamwork as a multi-level phenomenon involving team performance, interpersonal interactions, and individual behavioral traits.
To demonstrate the breadth of research opportunities enabled by these annotations, we formulate a set of benchmark prediction tasks spanning three levels of analysis.
At the \textbf{team level}, we consider overall teamwork assessment through OTAS as well as procedural phase duration prediction. These tasks capture collective team performance and workflow efficiency.
At the \textbf{individual level}, we evaluate behavioral and cognitive constructs derived from NOTSS and BFTQ together with leadership-oriented dimensions provided by SYMLOG and GLIS. These annotations allow the study of individual contributions to team functioning and decision-making.
At the \textbf{interaction level}, we consider communication modelling through the HMT framework, focusing on the structure and function of interpersonal exchanges.
To provide representative baselines, we evaluate the same families of models combined with the same set of enriched features introduced in the previous section, namely \SNN, \TNN, \ReNN, \TReNN, and \TEReNN. Across all tasks, performance is measured using macro F1-score to account for class imbalance. Results are averaged over ten random seeds and evaluated using a Leave-One-Group-Out (LOGO) protocol to assess generalization across unseen surgical teams.
Table~\ref{tab:combined_results} reports the results across all prediction tasks. Several observations emerge. First, meaningful predictive performance can be achieved across all annotation layers, suggesting that the proposed labels capture behavioral patterns that are consistently reflected in the multimodal observations. This finding demonstrates the learnability of the annotated constructs from multimodal observations and demonstrates the feasibility of learning a diverse range of teamwork-related constructs from the enriched representations.
Second, performance generally improves when models incorporate temporal and relational reasoning. While temporal architectures outperform static baselines on most tasks, relational and spatio-temporal models provide further gains, highlighting the importance of explicitly modelling interactions among team members.
Finally, tempo-relational architectures consistently achieve the strongest results across all levels of analysis. In particular, \TEReNN achieves the highest performance on most tasks, indicating that effective modelling of surgical teamwork requires jointly capturing behavioral evolution, interpersonal dependencies, and multimodal contextual information.
Overall, these benchmarks illustrate how the proposed dataset can support research ranging from team-level performance assessment to interaction analysis and individual behavior modelling, providing a unified evaluation framework for future multimodal studies of collaborative work in operating-room environments.



\subsection{Counterfactual case study}
\label{subsec:counterfactual_prediction}
\begin{figure}[t!]
    \centering
    \includegraphics[width=1\linewidth]{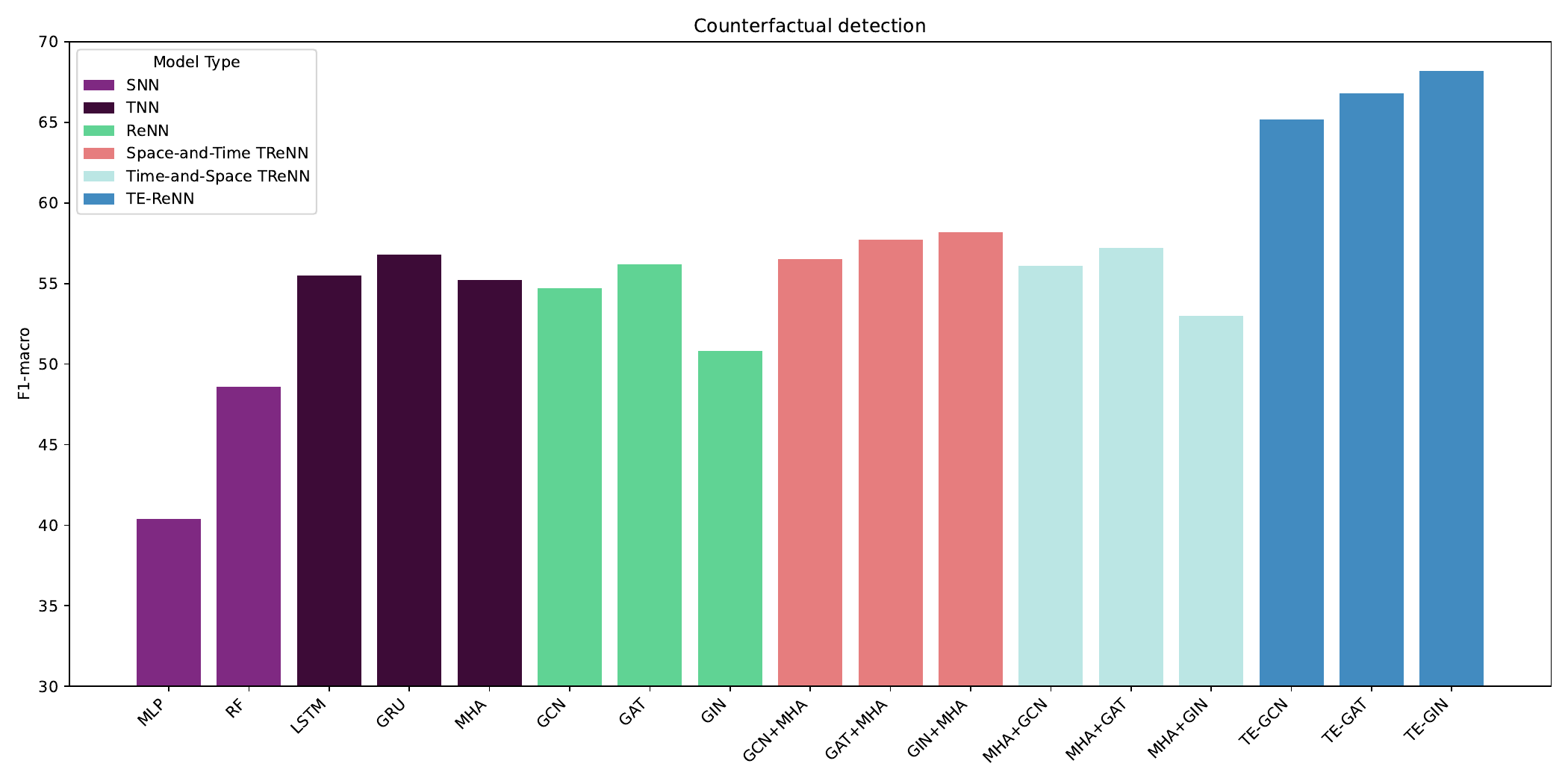}
    \caption{Test F1-macro computed according to the LOGO protocol for each team member, averaged across ten runs with different random seeds.}
    \label{fig:counterfactual_detection}
\end{figure}

The proposed dataset introduces a counterfactual annotation layer designed to capture situations in which teamwork could have been improved through a different behavior, decision, or interaction. To the best of our knowledge, this type of annotation is not available in existing operating-room benchmarks and opens new opportunities for studying teamwork failures, responsibility attribution, and explainable team modelling.
As a proof of concept, we formulate a benchmark task aimed at identifying counterfactual events directly from multimodal observations. More specifically, the task is cast as a binary classification problem in which each team member is classified as either being associated with a counterfactual event or not. The objective is not to determine causal responsibility, but rather to assess whether the multimodal signals captured by the dataset contain predictive patterns associated with situations that experts identified as potentially improvable.
To provide representative baselines, we evaluate the same modeling families adopted throughout the paper, namely \SNN, \TNN, \ReNN, \TReNN, and \TEReNN. All experiments are performed under the LOGO protocol and averaged across ten random seeds. Due to the rarity of counterfactual events, performance is evaluated using macro F1-score.
Figure~\ref{fig:counterfactual_detection} reports the results obtained by the different modeling paradigms. Despite the challenging nature of the task and the relatively limited number of annotated counterfactual events, all model families achieve performance substantially above chance level. This finding suggests that counterfactual situations are associated with observable behavioral and interactional patterns that can be captured through multimodal team representations.
Consistent with the trends observed in the previous benchmarks, performance generally improves as models incorporate richer temporal and relational inductive biases. Static models provide the weakest results, while temporal and graph-based approaches achieve progressively stronger performance. The best results are obtained by the \TEReNN family, indicating that recognizing potentially problematic situations requires jointly modeling behavioral evolution and interpersonal interactions.
More importantly, these results demonstrate the practical usefulness of the proposed counterfactual annotations. The benchmark shows that counterfactual events can be operationalized as learnable prediction targets, transforming qualitative expert reflections into computationally tractable tasks. This opens several future research directions, including early-warning systems for teamwork degradation, explainable assessment frameworks, and AI-assisted training tools capable of identifying situations in which alternative behaviors could have led to improved team performance.

\section{Discussion}\label{sec:discussion}

This work introduces a multimodal, interaction-centered extension of an operating room dataset designed to support the analysis of surgical teamwork through enriched temporal, relational, and socially grounded representations. By combining speaker-aware preprocessing, bilingual transcripts, multi-level teamwork annotations, and a counterfactual layer, the proposed resource moves beyond existing OR benchmarks focused primarily on perception or single-score evaluation.

Despite these contributions, several limitations should be acknowledged. First, the enrichment and annotation process relies heavily on expert and manual effort, which ensures high quality but limits scalability to larger datasets. Future work may explore semi-automated annotation pipelines to address this constraint. Second, the integration of multiple teamwork frameworks (e.g., OTAS, NOTSS, HMT, SYMLOG, GLIS, and BFTQ) preserves theoretical richness but introduces partial redundancy and potential inconsistencies across constructs. Rather than a drawback, this heterogeneity may also enable the study of cross-framework alignment and latent structure discovery.

Third, the counterfactual annotation layer provides useful explanatory signals about behaviors perceived to contribute to teamwork degradation, but remains inherently subjective, reflecting an annotator's interpretation rather than experimentally validated causal mechanisms. Nevertheless, it represents a first step toward more interpretable and explanation-aware modeling of team dynamics in high-stakes environments.

Finally, while the dataset is grounded in realistic OR recordings, privacy and accessibility constraints remain important considerations for broader dissemination and reproducibility. Addressing these challenges will be essential for scaling such resources in future work.

Overall, this work supports a shift from purely descriptive modeling of surgical workflows toward integrated, multimodal, and explainable modeling of teamwork, opening avenues for future research in temporal-relational learning, multi-task modeling, and human-centered AI for high-stakes collaborative settings.

\begin{acks}
Funded by the European Union. Views and opinions expressed are however those of the author(s) only and do not necessarily reflect those of the European Union or the European Health and Digital Executive Agency (HaDEA). Neither the European Union nor the granting authority can be held responsible for them. Grant Agreement no. 101120763 - TANGO. VMDL acknowledges the support of the MUR PNRR project FAIR - Future AI Research (PE00000013) funded by the NextGenerationEU. AL was supported by the Research Council of Norway through its Centre of Excellence Integreat - The Norwegian Centre for knowledge-driven machine learning, project number 332645. We acknowledge Raffaella Sabrina Fellone and Vincenza Moncelli for supporting the annotation process.
\end{acks}

\bibliographystyle{ACM-Reference-Format}
\bibliography{sample-base}

\end{document}